# Fast Learning and Prediction for Object Detection using Whitened CNN Features

Björn Barz    Erik Rodner    Christoph Käding    Joachim Denzler

Computer Vision Group
Friedrich Schiller University Jena
Ernst-Abbe-Platz 2
Jena, Germany

## Abstract

We combine features extracted from pre-trained convolutional neural networks (CNNs) with the fast, linear Exemplar-LDA classifier to get the advantages of both: the high detection performance of CNNs, automatic feature engineering, fast model learning from few training samples and efficient sliding-window detection.
The Adaptive Real-Time Object Detection System (ARTOS) has been refactored broadly to be used in combination with Caffe for the experimental studies reported in this work.

## 1. Introduction

In 2014, we presented the **A**daptive **R**eal-**T**ime **O**bject Detection **S**ystem (ARTOS) [1]. It features a fast method for learning linear classifiers from positive samples only, which are collected automatically from *ImageNet* [2]. The user may inspect a visualization of the model mixture, which is based on *Histograms of Oriented Gradients (HOG)* [3] as features, and can remove components from the model as well as add new ones learned from domain-specific in-situ images to improve the overall performance of the model for the task at hand. Object detection is performed by applying template-matching with a sliding-window approach, which is significantly sped up by leveraging the Fourier transform [4].

Though HOG features have previously been state-of-the-art for many years, an actually not-so-new method outperformed them recently: The huge amounts of available data gave new rise to the approach of *Neural Networks*, which originated in the 1980s. Nowadays, this method is used in the field of computer vision in the form of so-called *Convolutional Neural Networks (CNNs)*, which prepend a sequence of convolutional, activation and pooling layers to the actual fully-connected net [5]. The idea behind this architecture is that many feature extraction methods are based on convolving the image with a pre-defined filter followed by dividing the image into cells and pooling histograms or the average or maximum value from those cells to form the final feature vector. The eminent advantage of CNNs is that not only the weights for the classifier (i.e., the fully-connected net), but also the weights of those convolutions are learned automatically to minimize a specific loss function. This relieves researchers from the burden of manual feature engineering and leads to convolutional image features that are tuned towards the given task and data.

This research was supported by grant DE 735/10-1 of the German Research Foundation (DFG).

Yielding very promising results for the task of image classification, attempts to adapt CNNs for object detection have been made soon and led to regional classifiers [6]. These are performing better than detectors based on HOG features, but at a cost: First, a sliding-window approach for classifying every possible bounding box on multiple scales of the image with a CNN cannot be accelerated using the Fourier transform as it is possible with linear classifiers. Second, CNNs are usually very prone to overfitting and, thus, require a lot of diverse training data for every class.

By extending ARTOS, we aim for combining the best of both worlds: the automatic feature engineering of CNNs with a linear classifier which allows for both training and detection at high speed.

## 2. Refactoring of ARTOS

### 2.1 Generalization

The first version of ARTOS relied strongly on the FFLD library [4] for detection and HOG feature extraction. We have loosened this dependence and refactored ARTOS completely to work with arbitrary feature representations.[1]

Thanks to these efforts, it also became possible to switch feature extractors at run-time or to use several different feature extractors at once. In consequence, the type of features used for model creation and their hyper-parameters are now stored together with the learned parameters in the model file and, so that they can be restored for detection automatically.

### 2.2 Model Evaluation API and GUI

The first version of the ARTOS library already contained a `ModelEvaluator` class, which could be used to evaluate the performance of models on test data in terms of precision, recall, F-measure and average precision. But that class was mainly used by ARTOS internally for threshold optimization and was exclusive to the C++ library.

In ARTOS v2, this functionality has not only been exposed to the C interface of the library, but is by now also available for the user via the Python GUI for comfortable and easy model evaluation. Figure 1 shows the evaluation dialog where the user may specify the test data to be used. After the detector has been run for all selected models on all test images, a dialog will display several performance metrics as well as a recall-precision curve.

## 3. Extraction of CNN Features

The idea behind the project described in this report is to boost the object detection performance of ARTOS by replacing HOG features with the same features a CNN would use. That is, the output of the last layer before the fully-connected network, which handles the actual classification. Those features serve as input for the linear classifier learned by ARTOS using the Exemplar-LDA method [7]. A linear model for the CNN features will allow for fast template-matching and hopefully give better results than HOG models, since the features have been optimized for the distinction between the classes in question.

Fortunately, several CNNs pre-trained on *ImageNet* images from 1000 classes do already exist. We use the *Caffe* library [8] to propagate images through such nets and extract features from a layer of the network. ARTOS will automatically analyze the parameters of all layers from the input layer to the one used for feature extraction and derive the number of pixels in a cell and the number of pixels removed from the borders of the image based on the `stride`, `pad` and `kernel_size` parameters of convolutional and pooling layers.

---

[1] ARTOS v2 is available as open source on GitHub: https://github.com/cvjena/artos

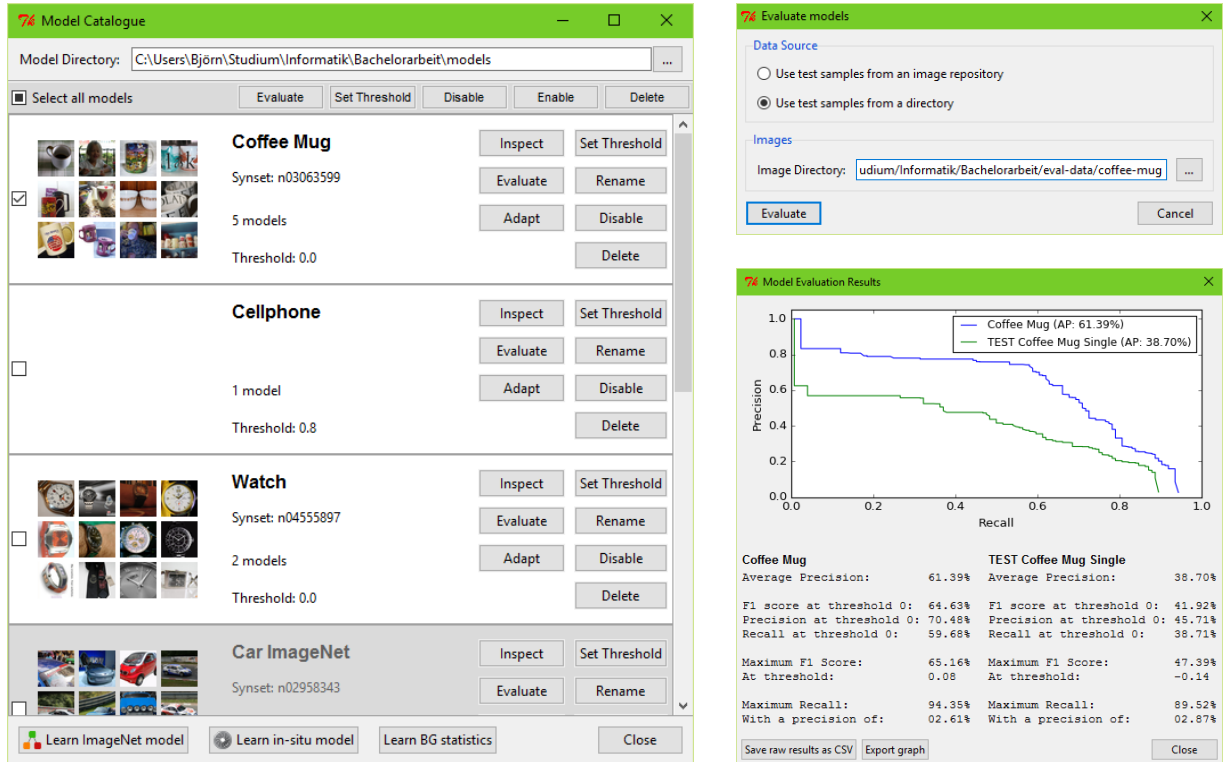

**Figure 1: Left:** New version of the model catalogue with selection controls for managing and evaluating multiple models at once. **Right top:** Dialog for selecting test images for evaluation. **Right bottom:** Evaluation results.

### 3.1 Pre- and post-processing

Most pre-trained networks have been learned on images from which the mean pixel value of each channel has been subtracted. Therefore, ARTOS must pre-process images in the same way.

Due to the high memory consumption of the internal data structures of Caffe and for higher throughput, we resize large images so that the larger dimension is not greater than 1024 pixels.

First experiments with CNN features ended in numerical problems, since the outputs of some CNN layers may be numerically large and the learning method of ARTOS needs to compute a covariance matrix of those features. In the end, most of these covariance matrices were practically asymmetric and not invertible. Scaling the features extracted from the CNN to the range $[-1,1]$ did not only solve this problem, but also improved the performance of models learned with features from channels which were not problematic before. But since doing so requires knowledge of the maximum possible value for each channel, an additional learning step had to be introduced for learning those maxima from some training images. ARTOS provides a separate tool for doing this and saves the detected maxima in a text file.

After all, the following steps are performed to extract CNN features from a given image:

1. Scale down the image if necessary.
2. Convert the image to the color space of the network (ARTOS uses RGB, while Caffe stores color images as BGR or may even expect grayscale images).
3. Subtract the mean pixel value for each channel from the image.
4. Resize the input layer of the net to match the size of the image and propagate that change to all layers up to the one which features will be extracted from.

5. Propagate the net from the first layer to the one which features are to be extracted from.
6. Extract the output from the *blob* belonging to the layer in question.
7. Scale the extracted features to the range $[-1,1]$ (approximately).

Before these features can be used to learn a model for any class, the following steps have to be performed only once:

1. Use Caffe to train an adequate CNN or find a pre-trained one.
2. Learn the maximum values of each feature channel of the layer in question from various scales of a large number of sample images.
3. Learn the mean feature cell and the autocorrelation function of the scaled CNN features from various scales of a large number of sample images.

### 3.2 PCA

Convolutional layers of CNNs usually have a large number of channels (around 256-512). This does not only result in high memory consumption for storing the features of images, but may also turn out to be prohibitive for the Exemplar-LDA learning method, which has to reconstruct a covariance matrix for all features and cells. Consider the scenario of learning a new model and let $M, N \in \mathbb{N}$ be the width and the height of the model in cells, respectively, and $F$ the number of feature channels. For the Exemplar-LDA learning method, a covariance matrix with $(M \cdot N \cdot F)^2$ coefficients had to be reconstructed from the pre-trained autocorrelation function. That means that the memory consumption of that covariance matrix grows quadratic with the number of features. For example, consider a rather small model size of $12 \times 12$ cells and a number of $512$ features. In that case, more than 20 GB of RAM would be required just for storing the covariance matrix.

One possible, but very time-consuming solution would be to define and train a CNN with a smaller number of feature channels. An easier approach, however, is to perform a dimensionality reduction by applying PCA on the features extracted from the net. ARTOS uses this technique to reduce the $F$ dimensions of a feature cell $x \in \mathbb{R}^F$ extracted from the net and already scaled to $[-1,1]$ to a new feature cell $\hat{x} = A \cdot (x - m)$.

ARTOS also ships with a tool that can be used to learn $A$ and $m$ from a number of training images, which adds one more step to the pre-learning procedure mentioned in 3.1. The effects of dimensionality reduction will be evaluated in section 4.4, while section 5 mentions another way to work around this issue.

### 3.3 Combinations of multiple layers

Another common problem regards the accuracy of the localization of the objects detected in an image: To reduce the number of input units for the fully-connected network and to mitigate the impact of minor spatial variations, most CNNs end up with quite large pooling regions like, for example, $32 \times 32$ pixels, which is quite much compared to the $8 \times 8$ pixel cells of HOG. While that is unproblematic for image classification, it affects the spatial resolution of object detection and potentially leads to many displaced bounding boxes. This is particularly fatal for narrow objects, where a displacement of just one cell may significantly affect the area of the intersection of the detected region with the ground-truth bounding box.

Following the solution proposed by [6], ARTOS is able to extract features from multiple layers of the CNN. These are then concatenated to form the final feature matrix, which will have the size of the largest layer, i.e., the one with the smallest cell size and highest resolution.

| Class | Training Images | | | Test Images | |
|---|---|---|---|---|---|
| | **Synset** | **Samples** | **Part of ILSVRC** | **Images** | **Objects** |
| **Airplane** | n02691156 | 376 | No | 670 | 865 |
| **Bottle** | n02876657 | 342 | Yes | 706 | 1259 |
| **Car** | n02958343 | 894 | Yes | 1161 | 2017 |
| **Chair** | n03001627 | 228 | Yes | 1119 | 2354 |
| **Cow** | n01887787 | 411 | No | 303 | 588 |
| **Person** | n00007846 | 178 | No | 4087 | 8566 |

**Table 1:** List of classes and datasets used for training and evaluation.

This requires oversampling of the smaller layers, which is done with nearest-neighbor interpolation. Since the resulting feature space will have a high dimensionality, PCA should be applied afterwards to reduce it.

The motivation of this approach is the idea, that the features of the deeper layers may be more useful for determining the class of a given region in the image. However, since those features have a rather low resolution, the features from upper layers may provide additional information that could support accurate localization.

We evaluate this approach in section 3.3.

## 4. Experiments and Evaluation

### 4.1 Datasets and CNNs

We have tested features extracted from several layers of two pre-trained CNNs: The first one is the BVLC reference net[2] (alias *CaffeNet*), which ships with *Caffe* and contains 5 convolutional layers, whereby the last such layer yields 256 feature channels. The second one is the VGG16 network[3] with 13 convolutional layers [9], which won the ILSVRC 2014 competition [10]. The last convolutional layer of that net yields 512 channels. Both nets have been trained on the 1000 classes of the aforementioned challenge.

The training images for the models learned by ARTOS come from *ImageNet* too, while we evaluate the performance of the models on images from PASCAL VOC 2012 [11] (*trainval* dataset). Experiments are made on six classes: airplane, bottle, car, chair, cow and person. Table 1 contains detailed information about the number of images used for training und evaluation per class.

### 4.2 The problem of nested detections

First inspections of qualitative detection results revealed a new problem arising from the use of CNN features: They perform just too well for the task of classification, which leads to fuzzy localization. It can be observed that a CNN is able to correctly classify an image if the object of interest is located in a sub-region of the image and does not fill it entirely. Likewise, a sliding-window detector using CNN features often does not only classify the smallest possible bounding box around an object as positive, but also a lot of larger bounding boxes containing that one. This results in nested detections with many of these being too large for being eliminated by non-maxima suppression. Some examples are shown in Figure 2. The detected bounding boxes often even take up almost the entire image.

---
[2] https://github.com/BVLC/caffe/tree/master/models/bvlc_reference_caffenet
[3] https://gist.github.com/ksimonyan/211839e770f7b538e2d8

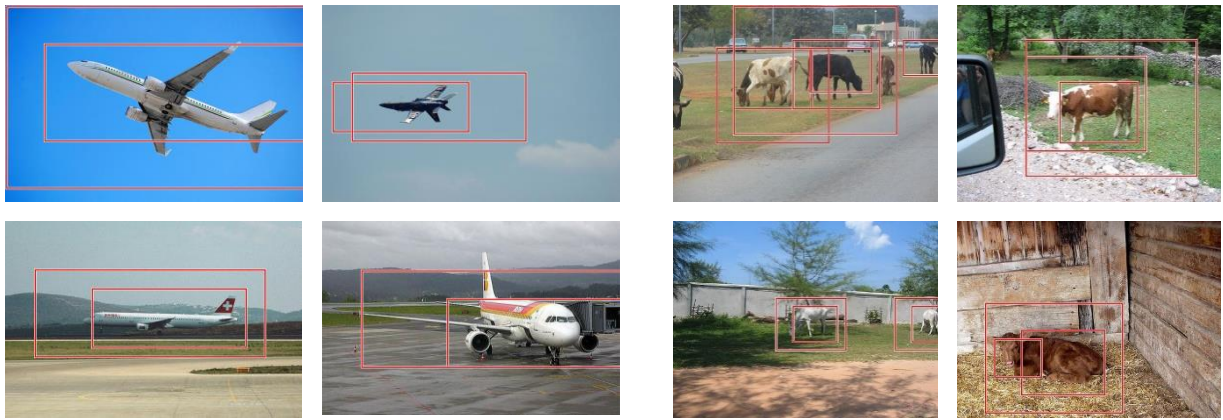

**Figure 2:** Examples of nested detections for the classes *airplane* (left) and *cow* (right).

A first attempt to deal with this problem by reducing the overlap threshold for non-maxima suppression from 0.5 to 0.4 increased the average precision of the detection results slightly, but did not eliminate the problem completely. We also investigated an approach for using separate thresholds for nested and for truly overlapping detections, but the impact on average precision was extremely small.

Thus, all results presented in the following for CNN features have been attained with a reduced overlap threshold of 0.4 for non-maxima suppression. The overlap threshold used for telling apart true and false positives during evaluation, which usually is 0.5, has not been altered, of course.

### 4.3 Performance of CNN features

First, we evaluated the performance of features extracted from single layers of a CNN and compared it to the average precision scored by a HOG model on the same dataset. The background statistics needed by the LDA learning method have been learned on 6000 images from ImageNet. This number of images is a trade-off between the quality of the statistics and the time needed for learning it. For practical purposes, one should rather use about 32000 images for learning statistics.

The results of these experiments as shown in Table 2 lead to several findings: There is no point in using features extracted from a pooling layer, since those contain less information than the layer they pool from and have a lower resolution. This motivated the idea, that features from convolutional layers may be more useful than features from activation layers, but this is clearly untrue for CaffeNet, where performance drops significantly when using the convolutional layer before the ReLU layer. The reason for this may be, that negative values of neurons in a convolutional layer carry no information for the network, since they will be set to 0 in the ReLU layer. Thus, only the absolute value of positive neurons is relevant, while the amplitude of negative values may be just noise. Interestingly, the situation is completely different for the VGG network, where features from the convolutional layer perform better than those from the activation layer in 3 out of 5 cases. It seems to depend on the network architecture and perhaps the training algorithm whether features from convolutional layers in front of an activation layer are useful or not.

Fast Object Detection using Whitened CNN Features

| AP | HOG | BVLC Reference Caffenet | | | VGG (16 layers) | | |
|---|---|---|---|---|---|---|---|
| | | pool5 | relu5 | conv5 | pool5 | relu5_3 | conv5_3 |
| **Person** | 4.37% | 22.98% | 25.23% | 21.63% | 24.52% | | |
| **Car** | 22.44% | 19.37% | 23.52% | 15.25% | 23.15% | 29.33% | 31.09% |
| **Chair** | 4.23% | 9.24% | 10.18% | 2.84% | 10.46% | 15.56% | 9.11% |
| **Cow** | 14.88% | 12.21% | 26.36% | 17.28% | 13.00% | 26.91% | 29.10% |
| **Bottle** | 15.41% | 12.47% | 14.97% | 6.72% | 14.15% | 20.27% | 20.01% |
| **Airplane** | 26.77% | 37.28% | 40.37% | 29.55% | 36.18% | 42.73% | 49.46% |
| **Model Size** | | | | | | | |
| **Person** | 9x12 | 5x7 | 11x14 | 11x14 | 5x7 | 11x14 | 11x14 |
| **Car** | 14x8 | 6x3 | 12x6 | 12x6 | 6x3 | 12x6 | 12x6 |
| **Chair** | 9x12 | 4x5 | 7x10 | 7x10 | 4x5 | 7x10 | 7x10 |
| **Cow** | 10x8 | 3x2 | 5x4 | 5x4 | 3x2 | 5x4 | 5x4 |
| **Bottle** | 6x18 | 2x7 | 5x14 | 5x14 | 2x7 | 5x14 | 5x14 |
| **Airplane** | 17x6 | 7x3 | 14x5 | 14x5 | 7x3 | 14x5 | 14x5 |
| **Cell Size** | 8x8 | 32x32 | 16x16 | 16x16 | 32x32 | 16x16 | 16x16 |

**Table 2:** Comparison of the average precision scored by models using features extracted from different CNN layers with models using HOG features. The best result for each class is highlighted in green and the worst one in red. The size of each model and the resolution of the feature cells is shown in the tables below the test results.

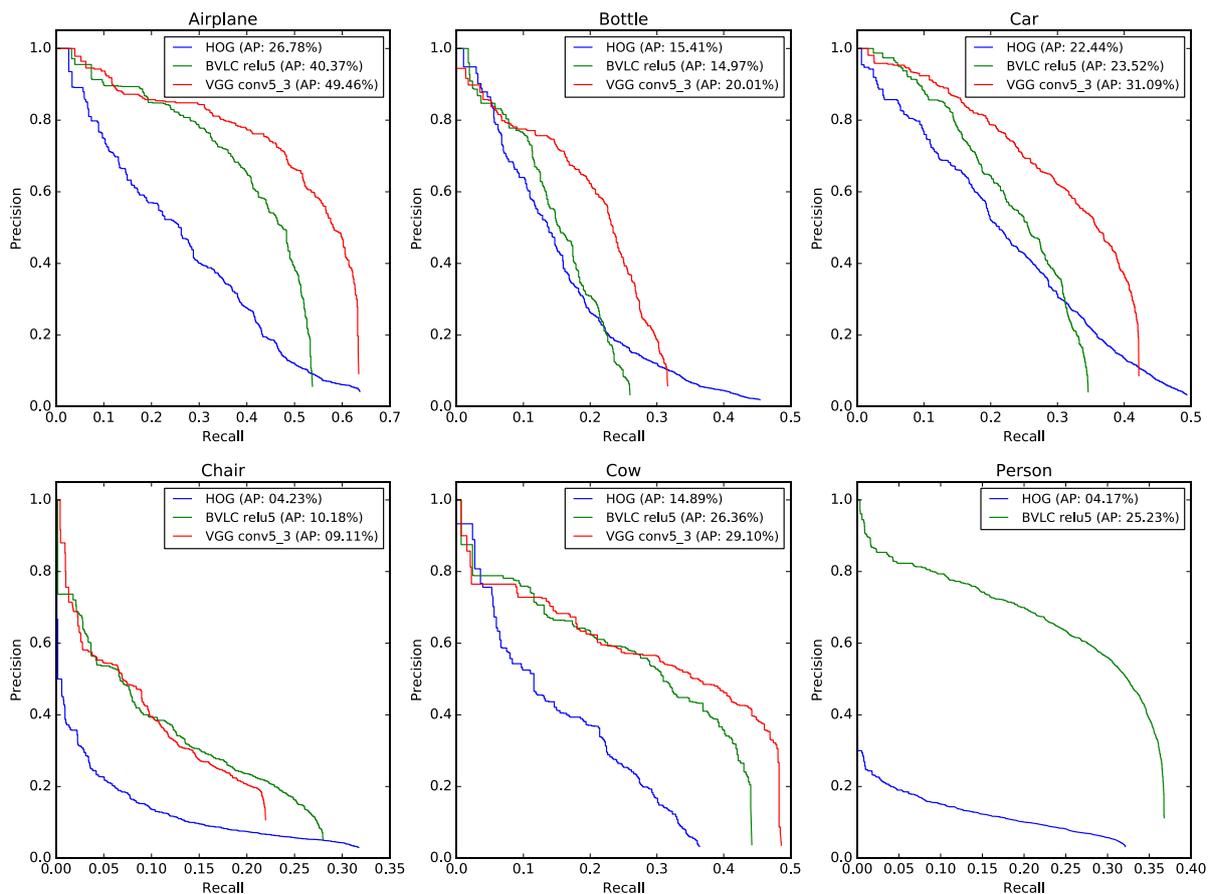

**Figure 3:** Recall-Precision graphs of models using features extracted from the best performing layer of each of the two CNNs compared to models based on HOG features.

Results for the class "person" are missing for the layers "relu5_3" and "conv5_3" of the VGG net, since they provided too much feature channels for a reconstruction of a covariance matrix of the respective model size. There was just not enough RAM. For the same reason, we were unable to test the performance of earlier layers with a smaller cell size.

While the features from CaffeNet are already superior to HOG features in almost all cases (with only one exception), the features from the much deeper VGG network clearly outperform HOG. This is also emphasized by the precision-recall graphs in Figure 3, where we compare the features from the best-performing layer of each of the two networks with HOG features.

In addition to this purely quantitative analysis, some qualitative examples and a roundup of common mistakes can be found in section 4.5.

It should be mentioned that previous works have found it beneficial to cluster training samples by aspect ratio as well as by HOG or Whitened HOG features and to learn a mixture consisting of models for each cluster [1, 7]. Figure 4 shows that this effect is a lot smaller when using CNN features. While clustering by aspect ratio may still result in a very small improvement, clustering by features does not take average precision any higher. This could indicate that the CNN features are already able to capture the characteristics of different sub-types of the same class, so that splitting up the training set in advance is not necessary. But even when clustering is applied for learning of HOG models, the models based on CNN features are still superior.

## 4.4 Impact of dimensionality reduction

One possible reason for the superiority of CNN features over HOG features might be the higher dimensionality of the feature space, which implies a higher amount of free parameters for the linear classifier. Therefore, we applied PCA for dimensionality reduction as described in section 3.2 and compared the performance of models using the reduced feature space with the performance of the models using the original features extracted from the CNN.

As shown in Figure 5, a reduction of the full feature space to 128 features can be performed without a significant loss of performance in almost all cases. Doing so may even result in better performance, since fewer free parameters leave smaller room for overfitting. Considering the class "person", dimensionality reduction is actually necessary for learning a model with features from the VGG net, since the use of 512 features led to too large covariance matrices for that class.

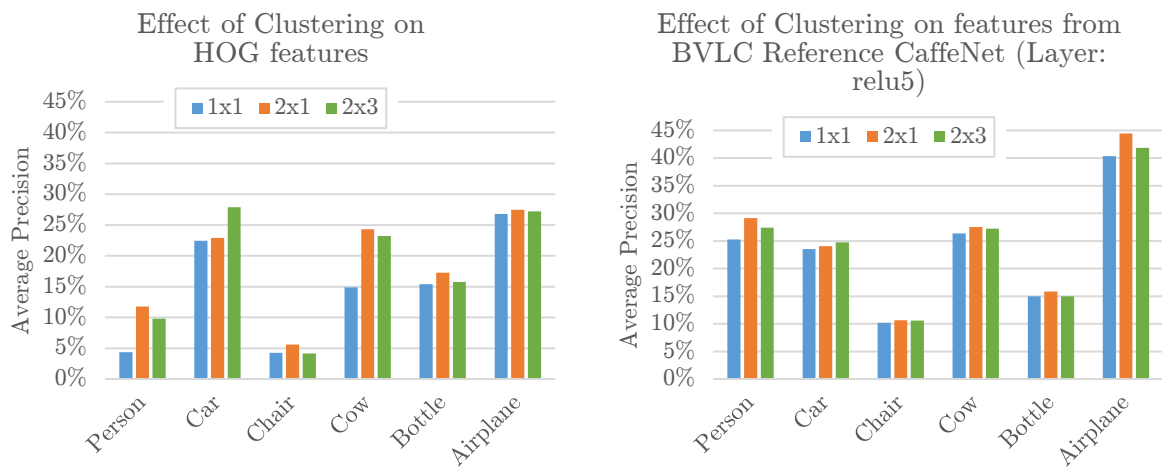

**Figure 4:** Comparison of performance of models using HOG features and models using CNN features under different clustering settings: no clustering (1x1), 2 aspect ratio clusters (2x1), 2 aspect ratio and 3 feature clusters per aspect ratio cluster (2x3).

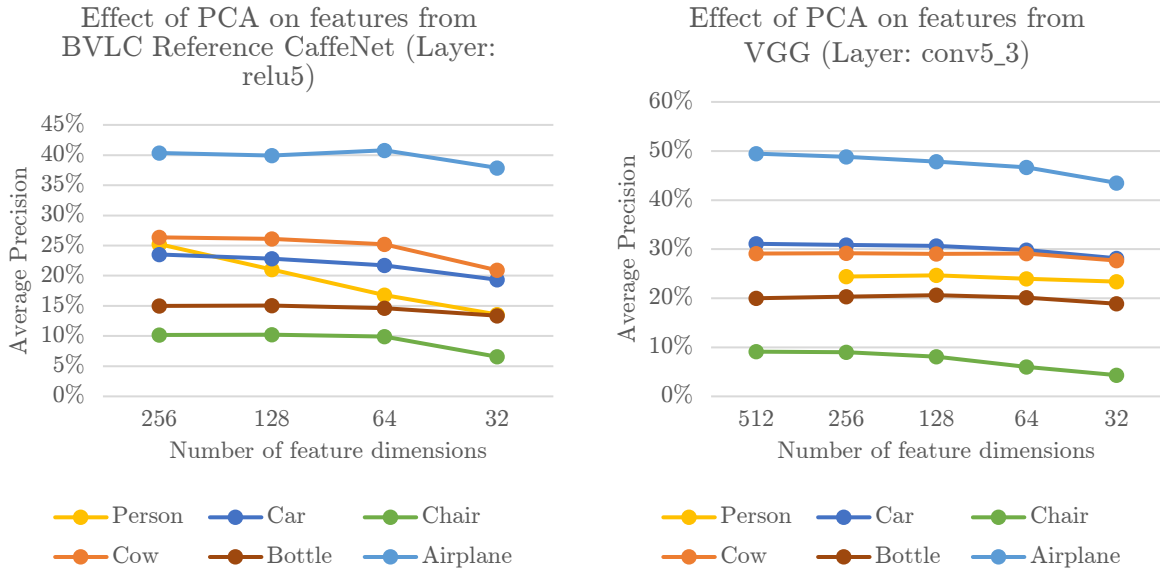

**Figure 5:** Performance of models using CNN features depending on the number of features.

While a reduction to 64 dimensions may also be reasonable given the improvements in speed and memory consumption implied by fewer dimensions, performance drops significantly when using only 32 features. Nevertheless, that case is of particular interest because that is also the number of features used by HOG and, thus, allows for a fair comparison. Despite the decreased performance compared to the full CNN feature space, the models using only 32 CNN features extracted from the VGG net still perform a lot better than models using HOG features (refer to Table 2 for the average precision of HOG models). In the case of the features extracted from CaffeNet and reduced to 32 dimensions, the classes "car" and "bottle" are better recognized by HOG models.

### 4.5 Qualitative results and error analysis

To get deeper insights into the performance and the problems of the CNN models than the flat average precision values provide, we have conducted further analyses for the features from the layer "conv5_3" of the VGG net using the VOC Error Diagnosis tool[4] [12].

The area plots in Figure 7 show the distribution of false positive types plotted against the number $N^* = N/N_j$ of top detections considered, where the number of detections $N$ is normalized by the total number of objects $N_j$ for the respective category $j$. That means, an optimal detector would yield 100% truly positive detections (white area) in the range [0,1]. A false positive is considered a localization error (blue area) if it fulfills the relaxed overlap criterion with threshold 0.1 instead of 0.5. Furthermore, the plots differentiate between confusions with similar categories (red area), non-similar objects (green area) and with the background (violet area). The recall under the usual overlap criterion is plotted as solid red line and the recall under the relaxed overlap criterion as dashed red line.

The detections for the class "cow" are mainly dominated by confusions with similar objects (76%), primarily sheep, horses and dogs (see also Figure 6). In consideration of the other four classes, localization errors take up between 15% and 21% of the top-ranked $N_j$ detections.

---

[4] http://dhoiem.cs.illinois.edu/projects/detectionAnalysis/

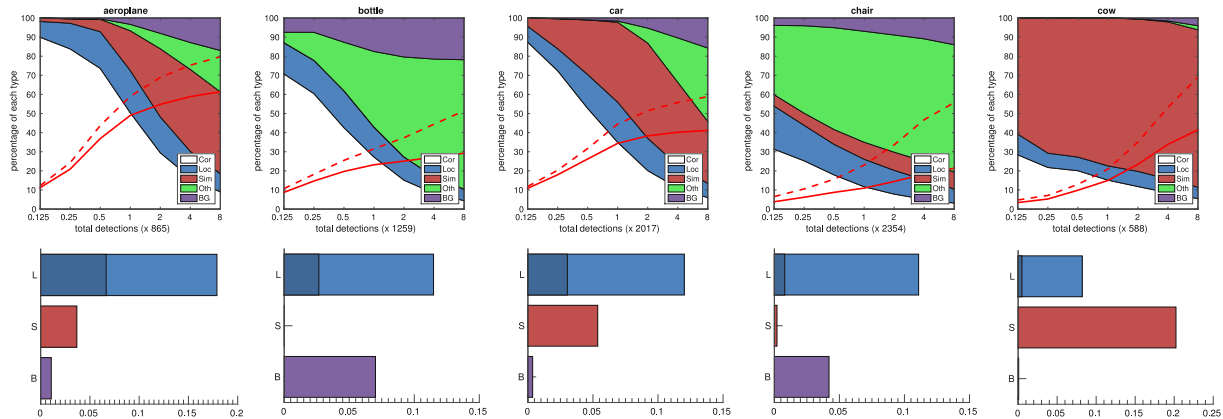

**Figure 7:** Distribution of false positive types (top) and impact of elimination of specific types on average precision (bottom) for models VGG features (layer conv5_3). See text for an explanation of these plots.

### Nested Detections

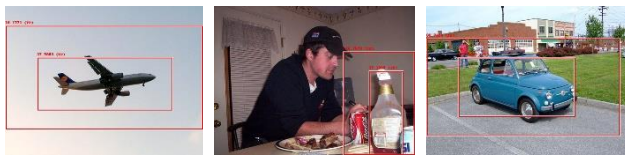

### Partial Detections

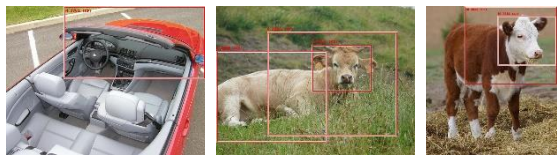

### Grouped Detections

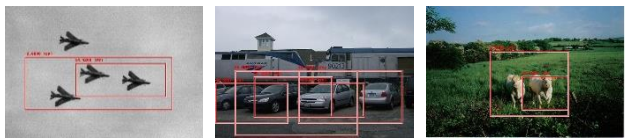

### Inaccurate Localization

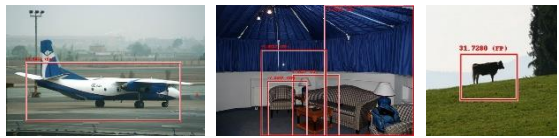

### Confusion with Similar Objects

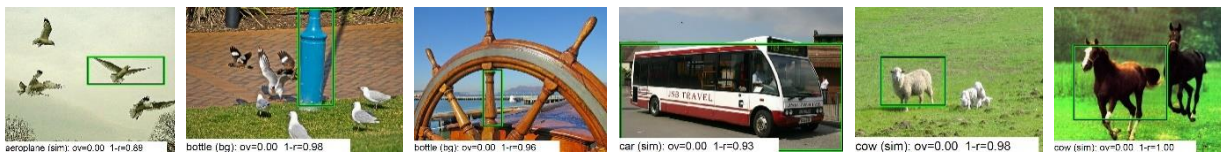

**Figure 6:** Examples of common types of false positives detected by a model using features from VGG 16 (layer: conv5_3) on Images from PASCAL VOC 2012 (trainval).

The bar plots below the area plots in Figure 7 visualize the possible impact on average precision if a specific type of false positives would be eliminated. For example, a better distinction between cows and similar animals could lead to an absolute improvement in terms of average precision by 0.2. In general, improving localization accuracy would result in large gains for all classes. The corresponding bar in the chart consists of two segments: The first one shows the impact if all duplicate or poor localizations would be removed, while the second one assumes that those detections would be corrected so that false positives would be turned into true positives, which is very effective. Doing so could improve average precision by between 0.09 and 0.17.

Figure 6 shows some examples of the most common types of errors observed during our experiments. The problem of nested detections, which belongs to the group of localization errors, has already been described in section 4.2. Besides inaccurate localization caused by, e.g., the coarse resolution of CNN features, another common problem of CNN features is the detection of multiple neighboring objects in a single bounding box. In contrast, often only parts are detected, in extreme cases even multiple parts of the same object, but not the object as a whole.

A qualitative comparison between detection performance of HOG models and models using features from the layer "conv5_3" of the VGG net is shown in Figure 9. For a threshold-neutral comparison, we have chosen the threshold for each model so that it maximizes the F1-score of the model on the data set. That means, the detections shown on the images are the best trade-off between precision and recall one can achieve with those models. The images shown in the figure have not been cherry-picked, but selected at random from the entire data set.

## 4.6 Combinations of multiple layers

Section 3.3 motivated extracting features from multiple layers of the same net for improved localization accuracy. Though [6] were successful with such an approach, corresponding experiments conducted using ARTOS with a linear classifier have been rather disappointing.

Table 3 compares the average precision of three multi-layer models with HOG models and the best single-layer model for each net. The features extracted from multiple layers have been transformed into a feature space with 128 dimensions (see 3.2). Though two of the multi-layer models have a higher or equal resolution as HOG, they perform even worse than HOG models in almost all cases. A more detailed analysis with the VOC Detection Analysis tool shown in Figure 8 reveals that those models have a very poor precision compared to Figure 7, though the amount of localization errors decreased slightly for some categories.

Further investigations would be necessary to figure out the cause of this poor performance. It may be related to the Exemplar-LDA learning method or its implementation. During the training of the models using features from three layers of the VGG network, we noticed that the reconstructed covariance matrices were not positive definite, so that a regularizer had to be added to the main diagonal of the matrix for a successful Cholesky decomposition.

| AP | HOG | BVLC (relu5) | VGG (conv5_3) | BVLC (relu2 + relu5 -> 128) | BVLC (relu3 + relu4 + relu5 -> 128) | VGG (conv3_3 + conv4_3 + conv5_3 -> 128) |
|---|---|---|---|---|---|---|
| Person | 4.37% | 25.23% | | 5.55% | 12.50% | 14.54% |
| Car | 22.44% | 23.52% | 31.09% | 11.14% | 8.91% | 12.90% |
| Chair | 4.23% | 10.18% | 9.11% | 1.94% | 2.13% | 2.32% |
| Cow | 14.88% | 26.36% | 29.10% | 7.70% | 13.03% | 17.09% |
| Bottle | 15.41% | 14.97% | 20.01% | 2.21% | 3.42% | 6.14% |
| Airplane | 26.77% | 40.37% | 49.46% | 21.27% | 26.98% | 19.64% |
| **Model Size** | | | | | | |
| Person | 9x12 | 11x14 | 11x14 | 15x20 | 11x14 | 15x20 |
| Car | 14x8 | 12x6 | 12x6 | 20x11 | 12x6 | 20x11 |
| Chair | 9x12 | 7x10 | 7x10 | 15x20 | 7x10 | 15x20 |
| Cow | 10x8 | 5x4 | 5x4 | 10x8 | 5x4 | 20x15 |
| Bottle | 6x18 | 5x14 | 5x14 | 7x20 | 5x14 | 7x20 |
| Airplane | 17x6 | 14x5 | 14x5 | 20x8 | 14x5 | 20x8 |
| **Cell Size** | 8x8 | 16x16 | 16x16 | 8x8 | 16x16 | 4x4 |

**Table 3:** Comparison of average precision scored by models using features extracted from multiple CNN layers compared to HOG and single-layer models. Best results per row highlighted in green, worst in red.

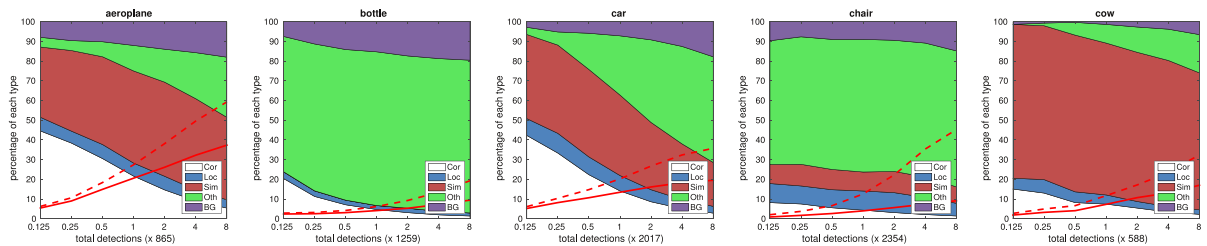

**Figure 8:** Distribution of false positive types for models using combined features from three layers of the VGG net: conv3_3, conv4_3, conv5_3. See section 4.5 for an explanation of these plots.

Fast Object Detection using Whitened CNN Features

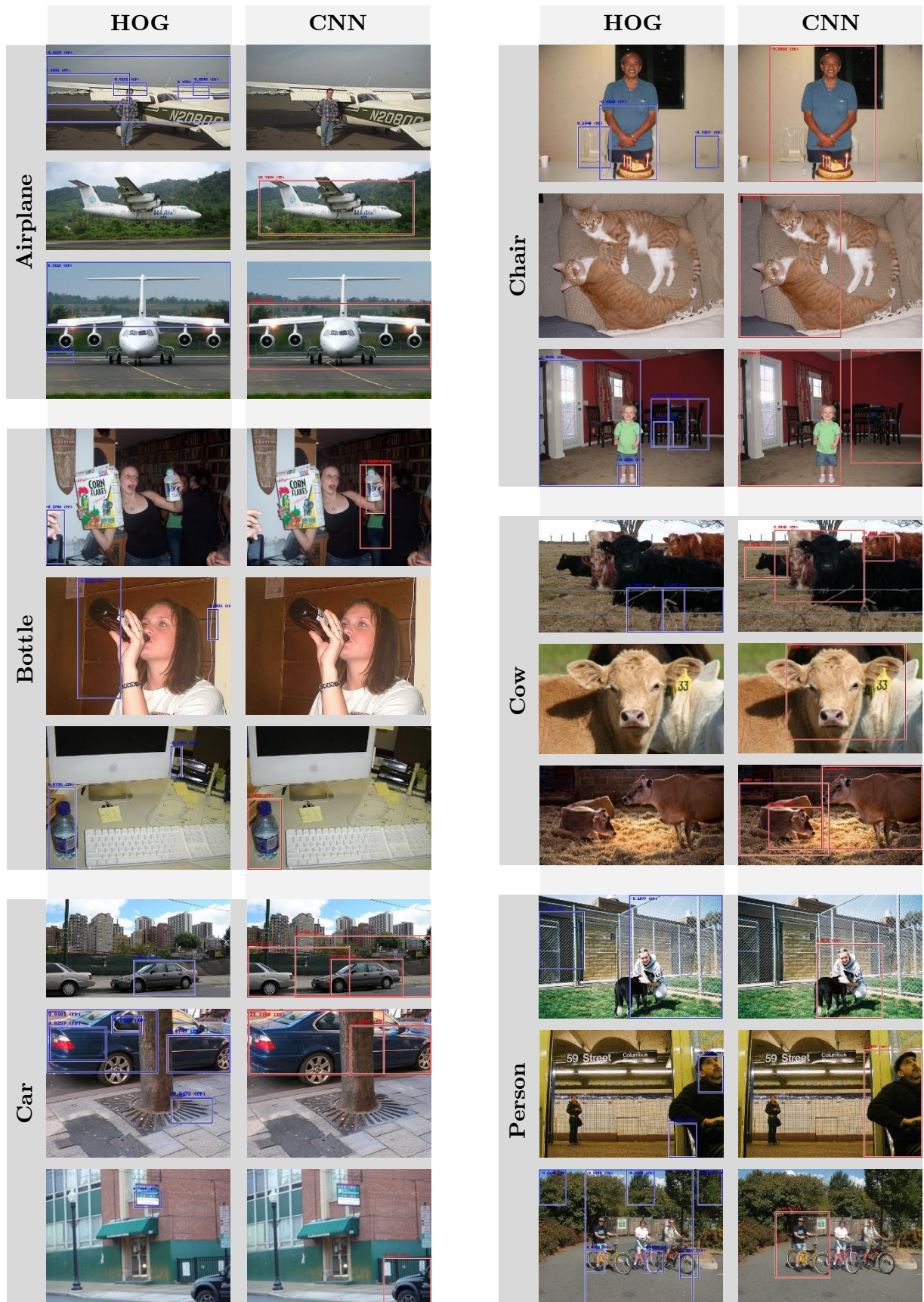

**Figure 9:** Exemplary images from the 6 VOC data sets with detections yielded by HOG models (blue) and models using features from the VGG net (layer conv5_3) (red).

## 4.7 Speed

The improved recognition performance of CNN features comes at the cost of a slower learning and detection speed. We have conducted the experiments described in this report on a machine with an Intel Core i7-4770 CPU (3.4 GHz quad-core) and an Nvidia Tesla K40c GPU.

While features can be extracted from the rather small CaffeNet without any overhead compared to the computation of HOG features, feature extraction from the more powerful VGG net takes six times as long. The major portion of time is spent forwarding the net, as illustrated in Figure 10 (left). However, when extracting features from multiple layers of different size, the extraction of the features from the memory of the net takes an even higher amount of time due to the necessary oversampling of the smaller layers.

But the speed of feature extraction is not the only factor that influences the speed of training and detection: the dimensionality of the feature space plays another important role. The size of the covariance matrix that must be reconstructed and decomposed for model learning grows quadratic with the number of features (see section 3.2) and dominates the entire learning process (see Figure 10 right). The Cholesky decomposition of that covariance matrix alone takes up 13 minutes for a model of size 10x8 cells when using 512 features.

Concerning detection, one Fourier transform must be performed for each feature plane. While ARTOS processes 4 images of size 640×480 pixels per second with HOG features, this throughput decreases to 0.33 images per second when using features from the layer "conv5_3" of the VGG network, which provides 512 features. In this case, the entire detection takes ten times as long as feature extraction. Thus, applying PCA for dimensionality reduction could be of great advantage for retaining real-time detection capabilities (see section 3.2).

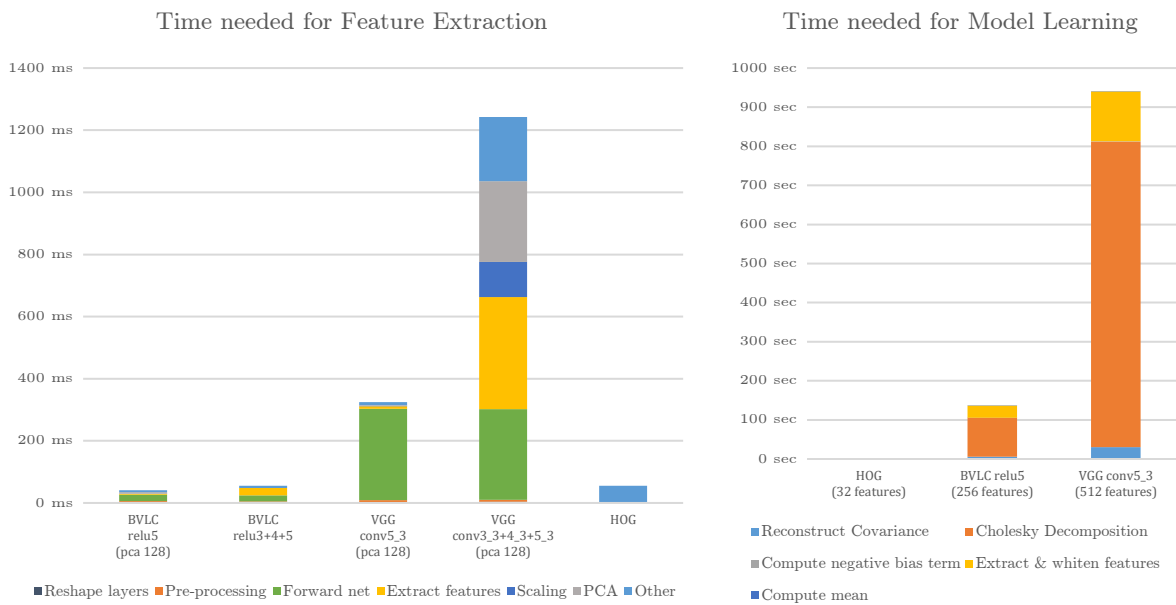

**Figure 10: Left:** Time taken by the single parts of the feature extraction procedure for an image of size 1024x786 pixels and different CNNs. **Right:** Time taken by the single components of the Exemplar-LDA model learning procedure for a model of size 10x8 cells and 302 training samples.

## 5. Shortcomings and Future Work

The refactoring of ARTOS described in section 2 made it possible to experiment with arbitrary features combined with the linear Exemplar-LDA learning method. In particular, features extracted from CNN networks have turned out to be clearly superior to HOG features. In contrast to existing CNN-based object detection algorithms, the use of a linear classifier allows for fast template matching by leveraging the Fourier Transform. But the advantages also come along with new problems, which have been described in section 4:

**Localization errors.** Most pre-trained CNNs end up with a rather large cell size, leading to a coarse localization resolution. While [6] solved this problem by combining features from multiple layers, this approach could not prove to be beneficial in combination with the ARTOS framework at all. Another possible solution mentioned by [6] could be to learn a bounding box regression separately, using the features from more fine-grained layers to predict a translation of the detected bounding box to improve localization accuracy. It may also be possible to use a separate classifier based on another type of features (from larger CNN layers or HOG features) to detect just the corners of the object in the neighborhood of the bounding box detected by the coarse-grained classifier. The predicted positions of the object corners could be used to adjust the position and size of the initial bounding box.

**Vague object boundaries.** In opposition to HOG features, CNN features seem to not depend that much on a strict alignment of the object boundaries with the bounds of the current window. As a consequence, nearby objects are often grouped in a single bounding box. On the other hand, nested bounding boxes around a single object can be observed quite often. A more sophisticated non-maxima suppression could help to eliminate these problems.

**Curse of dimensionality.** Typical CNNs usually provide hundreds of features. This does not only slow down detection, but may also preclude training of models due to the high amount of memory needed for storing the full covariance matrix. The dimensionality reduction experiments in section 4.4 show that such a high-dimensional feature space may be not necessary at all. But regardless of the possibility of dimensionality reduction, the covariance matrix reconstructed from the much more compact autocorrelation function is of a very special shape: It is a symmetric Block-Toeplitz-matrix, which is highly redundant, since blocks of coefficients are constant along any diagonal. Given this, the Cholesky decomposition might not be the most efficient method for solving the equation system that is the core of the WHO learning method. There may exists a method which does not need the full covariance matrix, but works on the autocorrelation function instead and, therefore, saves a lot of memory as well as time.

**Dedicated CNN architectures.** The CNNs used throughout this report for feature extraction have been learned in the face of non-linear neural network classifiers and the task of image classification instead of object detection. It would be interesting to see if net architectures tuned for the task of detection could yield better results. The last convolutional layer of such nets should contain a reasonable number of feature planes (less than 128) for the sake of speed, but with a rather small cell size (e.g. 8×8 pixels) for better localization accuracy. An even more consistent approach could be to already learn the coefficients of the convolutional layers with respect to loss information generated by a linear classifier instead of a large fully-connected neural net. A binary linear classifier could easily be modeled by a single fully-connected layer with a single neuron.